\documentclass[10pt,twocolumn,letterpaper]{article}

\usepackage{cvpr}              %

\usepackage{amsmath}
\usepackage{amssymb}
\usepackage{amsfonts}
\usepackage{mathabx}
\usepackage{tabularx}
\usepackage{diagbox}

\usepackage{siunitx}  %

\providecommand{\mat}[1]{\mathsf{#1}}

\usepackage{mathtools}
\usepackage{enumitem}

\newcommand{\ours}{$\mat{A}\cdot$LoMa}

\newcommand{\sunetal}{Sun et~al\onedot}
\newcommand{\sunetalcite}{\citet{sun2025learning}\xspace}
\newcommand{\sunetalname}{Sun et~al.}

\newcommand{\baffnet}{KN+HN+AN\xspace}     %
\newcommand{\bsesti}{SIFT+HN+S3E\xspace}   %
\newcommand{\blomaff}{LoMa+S3E+AN\xspace}  %
\newcommand{\bkeynet}{KN+HN\xspace}        %
\newcommand{\bsift}{SIFT+HN\xspace}        %
\newcommand{\bdkm}{DKM\xspace}             %
\newcommand{\solvsift}{UP1SIFT\xspace}  %
\newcommand{\solvpac}{P1AC\xspace}       %
\newcommand{\solvac}{UP1PfAC\xspace}      %

\definecolor{tabred}{rgb}{1, 0.7, 0.7}
\definecolor{taborange}{rgb}{1, 0.85, 0.7}
\definecolor{tabyellow}{rgb}{1, 1, 0.7}

\definecolor{tabgreen1}{HTML}{71AD8B}
\definecolor{tabgreen2}{HTML}{B6D6B5}
\definecolor{tabgreen3}{HTML}{E5EFE0}

\colorlet{firstcolor}{tabgreen1}
\colorlet{secondcolor}{tabgreen2}
\colorlet{thirdcolor}{tabgreen3}

\newcommand{\best}{\cellcolor{firstcolor}}
\newcommand{\better}{\cellcolor{secondcolor}}
\newcommand{\good}{\cellcolor{thirdcolor}}

\definecolor{refgraycol}{HTML}{8A8A8A}
\newcommand{\refrow}[1]{\textcolor{refgraycol}{#1}}

\usepackage{tcolorbox}

\usepackage{pifont}
\definecolor{cmarkgreen}{HTML}{2E8B57}
\definecolor{xmarkred}{HTML}{C0392B}
\newcommand{\cmark}{\textcolor{cmarkgreen}{\ding{51}}}
\newcommand{\xmark}{\textcolor{xmarkred}{\ding{55}}}
\definecolor{snowblue}{HTML}{5DADE2}
\newcommand{\snow}{\textcolor{snowblue}{\ding{102}}}
\usepackage{fontawesome5}
\definecolor{flameorange}{HTML}{E67E22}
\newcommand{\flame}{\textcolor{flameorange}{\scalebox{0.8}{\faFire}}}

\definecolor{cvprblue}{rgb}{0.21,0.49,0.74}
\usepackage[pagebackref,breaklinks,colorlinks,citecolor=cvprblue]{hyperref}

\usepackage{colortbl}
\usepackage{booktabs}
\usepackage{arydshln}
\newcommand{\dashedmidrule}{\noalign{\vskip\aboverulesep}\hdashline\noalign{\vskip\belowrulesep}}
\usepackage{multirow}
\usepackage{cuted}      %
\usepackage{afterpage}  %

\newcommand{\RN}[1]{\uppercase\expandafter{\romannumeral #1}}

\def\confName{CVPR}
\def\confYear{2027}

\title{\texorpdfstring{\ours}{A-LoMa}: Affine Correspondences Beyond Patches}

\author{
David Nordström$^{1}$ \hspace{2em} Georg Bökman$^{2}$ \hspace{2em} Viktor Larsson$^{3}$ \\
Fredrik Kahl$^{1}$ \hspace{2em} Jonathan Ventura$^{4}$ \\
{\small $^1$Chalmers University of Technology \hspace{2em} $^2$University of Amsterdam} \\
{\small $^3$Lund University \hspace{2em} $^4$California Polytechnic State University}
}

\begin{document}
\raggedbottom
\maketitle

\begin{strip}
\vspace{-20pt}
\centering
\includegraphics[width=\textwidth]{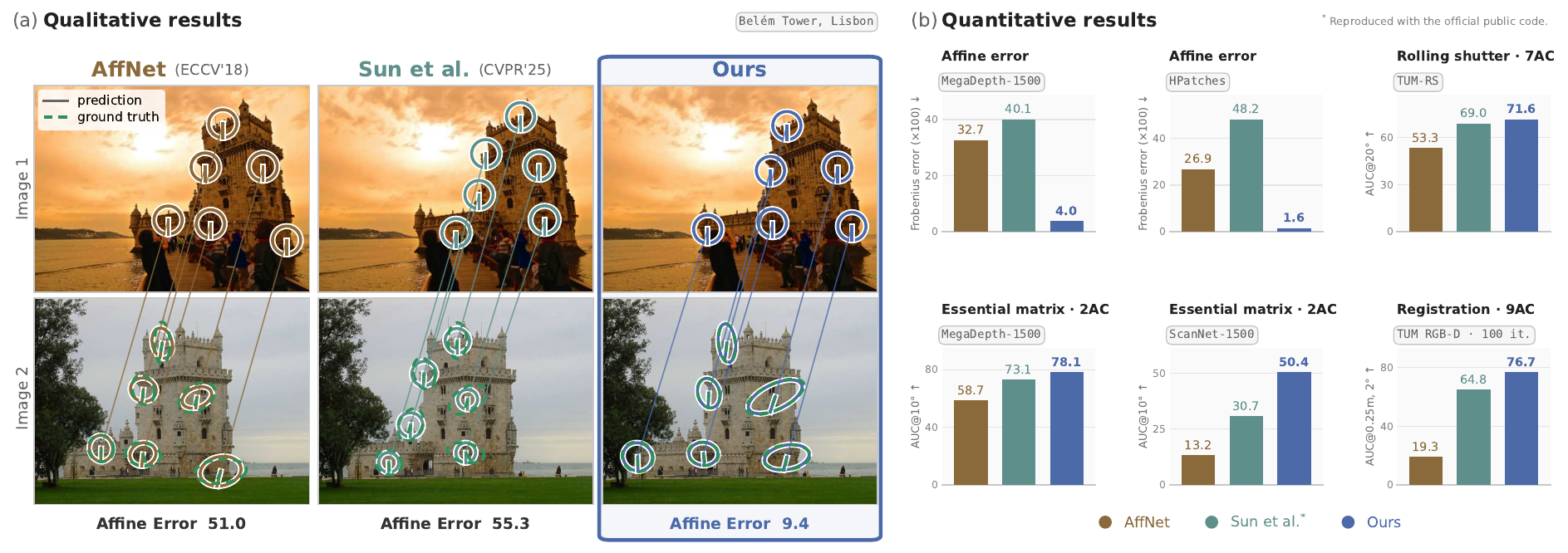}
\vspace{-6pt}
\captionsetup{type=figure,font=small}
\caption{\textbf{Accurate estimation of ACs.}
\textbf{(a)}~A circle at a keypoint in image~1 is mapped into image~2 by the AC predicted by KeyNet+HardNet+AffNet (AffNet), \citet{sun2025learning}, and \ours-B~(solid), versus ground truth (dashed green). \textbf{(b)}~Quantitative performance shows \ours-B reduces affine error and improves accuracy on downstream tasks like essential matrix estimation from 2 ACs.}
\label{fig:teaser}
\vspace{-8pt}
\end{strip}

\afterpage{\flushbottom}

\begin{abstract}
Affine correspondences (ACs) enable robust and efficient solvers for multiple-view geometry by constraining the local geometry of point correspondences (PCs). Despite their utility in minimal solvers, AC estimation remains understudied. We revisit AC estimation beyond local image patches, extending it to the image and image-pair levels. Rather than estimating the components of the affine transformation separately, we learn the full 4-DoF matrix jointly. Building on this formulation, we introduce \ours, an AC estimator using LoMa keypoint descriptions, demonstrating consistent improvements across a wide range of downstream tasks. Code at \href{https://github.com/davnords/loma}{https://github.com/davnords/loma}.
\end{abstract}
\section{Introduction}\label{section:introduction}

Geometric constraints are fundamental to multiple-view geometry~\cite{Hartley_Zisserman_2004}, with image correspondences providing the basis for many of its core estimation problems. \textit{Affine correspondences} (ACs) enrich point correspondences (PCs) with the local $2\times2$ affine transformation between the corresponding image regions, providing four additional degrees of freedom (DoF) per match. This additional information enables minimal solvers with far fewer correspondences: for example, relative pose estimation requires 5 PCs~\cite{nister2004efficient}, but only 2 ACs~\cite{Barth2018EfficientRO}. ACs have similarly been exploited for surface-normal estimation~\cite{hajder2023fast}, homography estimation~\cite{barath2017homography}, epipolar geometry~\cite{Hruby2024, Raposo_2016_CVPR, Barath2026RollingShutter, perdoch2006epipolar, Barth2018EfficientRO}, absolute pose~\cite{koser2008differential,Ventura_2023_ICCV}, and focal length estimation~\cite{barath2017minimal,valtonen-ornhag-jaenal-2026-eccv}.

Despite these advances in using ACs, estimating ACs themselves has received little attention. Traditional approaches rely on hand-crafted affine region detectors, such as Harris-Affine~\cite{mikolajczyk2002affine} and Hessian-Affine~\cite{mikolajczyk2004scale}, followed by iterative refinement such as Baumberg iteration~\cite{baumberg2000reliable}. Learned approaches later improved this pipeline, most notably AffNet~\cite{mishkin2018repeatability}. However, since its publication in 2018, progress in AC estimation has been limited, while PC estimation has seen many innovations.

Modern PC estimators reason over global context within and across views, rather than treating correspondences as independent local patches. Detector-free methods such as DKM~\cite{edstedt2023dkm} and RoMa~\cite{edstedt2024roma,edstedt2026romav2}, as well as learned sparse matchers such as SuperGlue~\cite{sarlin2020superglue} and LoMa~\cite{nordstrom2026lomalocalfeaturematching}, have substantially improved the quality and robustness of PCs. Yet AC estimation has largely remained disconnected from this progress. To our knowledge, \sunetalcite~is the only approach that combines a modern matcher with learned affine estimation. However, the affine transformation is still predicted from a small local patch around each match, and its four DoF are not predicted. Instead, one network estimates scale and orientation, and a separate one the affine shape.

This raises a natural question: \textit{can AC estimation benefit from the same contextual information that has improved PC estimation?} Existing approaches reduce each correspondence to a local patch, discarding the image-wide and cross-view information already available in modern matchers. Moreover, the four components of an AC are coupled, making their joint prediction challenging from local appearance alone. Indeed, previous work has reported difficulties with jointly estimating these components~\cite{mishkin2018repeatability,sun2025learning}, motivating decomposed predictions and multi-stage estimation.

In this work, we revisit AC estimation from a contextual perspective. Instead of treating each correspondence in isolation, we progressively incorporate image-level and cross-view information, and find that expanding the context from a patch to the full image pair sharply reduces affine error. Building on this, we augment a modern matcher with a lightweight affine head that directly predicts the full 4-DoF affine transformation. Joint prediction of scale, orientation, and shape then becomes both feasible and accurate.

Our resulting family of models, \ours, advances the state of the art in AC estimation with a simpler formulation. As shown in \cref{fig:teaser}, it reduces affine estimation error by an order of magnitude compared with AffNet and \sunetal, while improving downstream geometric estimation.

\noindent\textbf{The main contributions are as follows:}
\begin{enumerate}[label=(\roman*), topsep=0pt]
    \item \textbf{Context matters for affine estimation.} We study the effect of context on AC prediction and show that image-level and cross-view information substantially reduce affine estimation error (\cref{fig:model-context}).
    \item \textbf{Joint prediction of scale, orientation, and affine shape is feasible.} We show that a context-aware model can directly predict all four affine DoF jointly, avoiding the decomposed and multi-stage estimation used by prior approaches~\cite{mishkin2018repeatability,sun2025learning}.
    \item \textbf{An order of magnitude lower affine error.} \ours, trained directly with ground-truth (GT) affine transformations, achieves an order of magnitude lower affine error than AffNet and \sunetal, as visualized in \cref{fig:teaser}, while improving downstream geometric tasks, \eg, achieving higher rolling-shutter pose accuracy with 7 ACs than with 20 PCs.
\end{enumerate}

\begin{figure*}[t]
\centering
\includegraphics[width=0.99\linewidth]{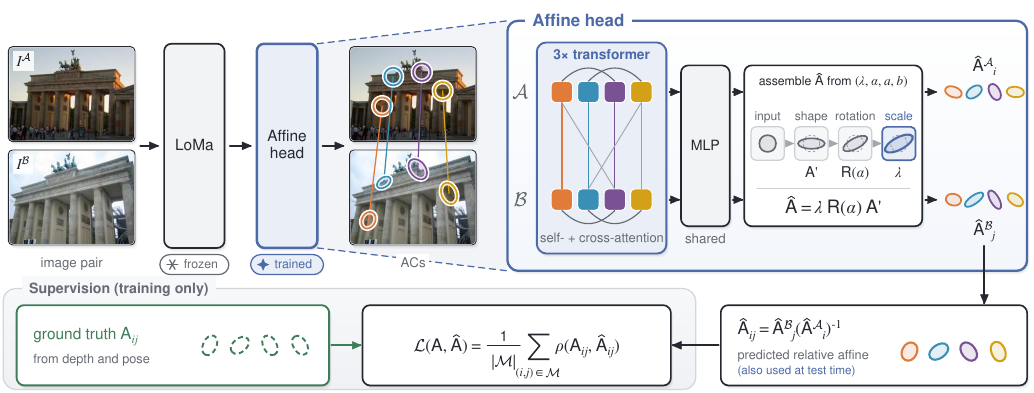}
\caption{\textbf{Architecture.} \ours~inputs an RGB image pair and uses LoMa to obtain PCs together with keypoint descriptors. The keypoint descriptors are fed into an affine head, implemented as a transformer. For each keypoint, a local affine frame is predicted. The frames of each match are composed into an affine transformation $\hat{\mat{A}}_{ij}$, which is supervised directly by the GT $\mat{A}_{ij}$ derived from depth maps.}
\label{fig:architecture}
\end{figure*}

\section{Background} \label{sec:related}

\subsection{Affine Correspondences}

An AC consists of a pair of matching keypoints together with the local $2\times2$ affine transformation mapping the neighborhood of one keypoint onto the neighborhood of the other.  Compared to a single PC, an AC provides additional geometric constraints~\cite{perdoch2006epipolar,bentolila2014conic} on the camera poses. As a consequence, minimal solvers require only 2 ACs (vs. 5 PCs) to estimate the essential matrix~\cite{Barth2018EfficientRO,barath2020making}, 3 ACs (vs. 7 PCs) for the fundamental matrix~\cite{Raposo_2016_CVPR,barath2020making}, and 2 ACs (vs. 4 PCs) for homographies~\cite{barath2016novelways,barath2019homographytwo}, which reduces to a single AC once the epipolar geometry is known~\cite{barath2017homography}. This has motivated their adoption in minimal solvers across many tasks, such as multi-camera and generalized motion~\cite{ventura2020multicamera,xu2025generalized,guan2026complete}, absolute pose estimation~\cite{Ventura_2023_ICCV,valtonen-ornhag-jaenal-2026-eccv,Ventura_2024_CVPR}, relative pose estimation~\cite{guan2020minimal,eichhardt2018affine,barath2022relative,yu2023globallyoptimal,Hruby2024,Barath2026RollingShutter}, planar and vehicle-mounted motion~\cite{hajder2020relativeplanar,valtonenornhag2024planar}, focal length estimation~\cite{barath2017minimal,yu2026minimalfocal}, and surface normal estimation~\cite{hajder2023fast}. More broadly, ACs are one of several ways to shrink the minimal sample. Alternatives include a known gravity direction~\cite{kukelova2010closed,sweeney2015efficient,valtonen-ornhag-jaenal-2026-eccv} and monocular or relative depth~\cite{Hruby2024,astermark2024fast,ding2024fundamental}. ACs have thus been incorporated into a wide range of minimal solvers. However, \textbf{estimating them has received far less attention.}

Traditionally, ACs were obtained from hand-crafted affine-covariant region detectors~\cite{mikolajczyk2005comparison, forssen2007shape, mikolajczyk2002affine, mikolajczyk2004scale}. AffNet~\cite{mishkin2018repeatability} proposed a learning-based method for estimating the affine shape and remains widely used almost a decade later. The network predicts an affine shape for each keypoint by decomposing the affine transformation as $\mat{A} = \lambda\, \mat{R}(\alpha)\, \mat{A}'$, where $\mat{A}'$ is a unit-determinant upper-triangular \emph{shape} matrix. An affine transformation $\mat{A}$ has four DoF: two from the shape ($\mat{A}'$), one from the scale ($\lambda$), and one from the orientation ($\mat{R}(\alpha)$). AffNet acts on local $32\times32$ image patches and is trained by optimizing the matching discriminability, applying synthetic affine perturbations with increasing difficulty through a curriculum. Practically, to estimate ACs using AffNet one combines its affine shape estimation with scale and orientation from separate networks. S3Esti~\cite{yan2022learning} is one such network, estimating only the affine scale and orientation of a keypoint, trained using a self-supervised probabilistic covariant loss that treats GT labels as latent variables. Similar to AffNet, S3Esti uses only local $32\times32$ patches around its keypoint for its affine estimate. \citet{dai2020boosting} also uses local patch crops but does so around a \emph{pair} of keypoints rather than a frame per keypoint.

\sunetalcite combine AffNet for shape (2 DoF) and S3Esti for scale and orientation (2 DoF), two separate CNNs, with the dense matcher DKM for extracting PCs. For supervision they use the Sampson distance of the affine epipolar constraint. The $32\times32$ patches cropped around the keypoints are the only source of information for the affine transformation, while a finetuned DKM is used for PCs.

Three common themes in prior work on learned AC estimation are reliance on \textbf{local patch information}, \textbf{indirect supervision}, and \textbf{separate networks for estimating scale, orientation, and shape}. We instead adopt a simpler, more effective formulation that uses \textbf{global information}, \textbf{direct supervision}, and estimates \textbf{all degrees of freedom jointly}.

\subsection{Point Correspondences}

For a long time, estimating PCs was closely tied to AC estimation.
SIFT~\cite{lowe2004distinctive} descriptors, for instance, are computed on patches normalized for orientation and scale, so this local geometry has to be estimated as part of the algorithm.
As learning-based approaches overtook PC estimation, the dominant idea instead became that affine invariance should be learned from data.
SuperPoint~\cite{detone2018superpoint} proposed a self-supervised protocol to learn to detect and describe keypoints, and SuperGlue~\cite{sarlin2020superglue} replaced the mutual-nearest-neighbor matcher with a graph neural network. Detector-free matching was popularized by works such as LoFTR~\cite{sun2021loftr}, DKM~\cite{edstedt2023dkm}, and RoMa~\cite{edstedt2024roma,edstedt2026romav2}, while further advances have been made for keypoint detection~\cite{barroso2019key, verdie2015tilde, edstedt2024dedodev2, edstedt2025dad, kim2024learning}, description~\cite{tian2019sosnet, germain2020s2dnet, edstedt2024dedode, bokman2024steerers}, joint detection and description~\cite{revaud2019r2d2, tyszkiewicz2020disk, Wang_2021_ICCV, zhao2022alike, Zhao2023ALIKED}, and matching~\cite{lindenberger2023lightglue,nordstrom2026whohandlesorientation,nordstrom2026lomalocalfeaturematching}. We build on the recent sparse matcher LoMa~\cite{nordstrom2026lomalocalfeaturematching} and extend it from PCs to ACs using its keypoint descriptors.

\begin{figure*}[t]
\centering
\includegraphics[width=0.99\linewidth]{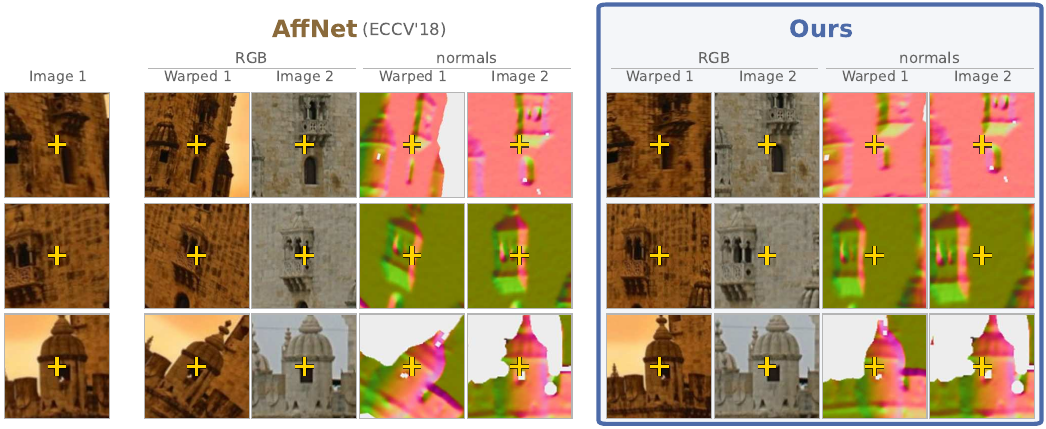}
\caption{\textbf{Local affine transformation.} For three correspondences, we warp the image patch around the keypoint (cross) with the predicted affine and compare it with the matched patch in the other image. Accurate estimates yield similar warped and matched patches.}
\label{fig:qualitative-patch}
\end{figure*}

\section{Method}

We fix notation (\cref{subsec:preliminaries}), then detail AC prediction (\cref{subsec:pred-acs}), the architecture (\cref{subsec:architecture}), and the learning objective (\cref{subsec:supervision}). See \cref{fig:architecture} for an overview of \ours.

\subsection{Notation}\label{subsec:preliminaries}

By $\mat{x}_i^{\mathcal A}$ and $\mat{x}_j^{\mathcal B}$ we denote a pair of corresponding keypoints in images $\mat{I}^{\mathcal A}$ and $\mat{I}^{\mathcal B}$. The two images are locally related by a warp $\phi: \Omega^{\mathcal A} \rightarrow \Omega^{\mathcal B}$ between their image domains, \ie, $\mat{x}_j^{\mathcal B} = \phi(\mat{x}_i^{\mathcal A})$, induced by the 3D scene and generally known only implicitly, \eg, through depth maps. The affine transformation of an AC (\cref{sec:related}) is then the local behavior of $\phi$ around $\mat{x}_i^{\mathcal A}$, given by its Jacobian $\mat{A}_{ij}\in \mathbb{R}^{2\times2}$,
\begin{equation}
    \phi\left(\mat{x}_i^{\mathcal A}+\mat{v}\right) \approx \mat{x}_j^{\mathcal B} + \mat{A}_{ij}\,\mat{v}, \qquad \mat{A}_{ij} := \left.\frac{\partial \phi}{\partial \mat{x}}\right|_{\mat{x}_i^{\mathcal A}},
    \label{eq:jacobian}
\end{equation}
for a small displacement $\mat{v}\in\mathbb R^2$, \ie, $\mat{A}_{ij}$ describes how $\mat{x}_j^{\mathcal B}$ moves under a small displacement of $\mat{x}_i^{\mathcal A}$.

\subsection{Predicting Affine Correspondences}\label{subsec:pred-acs}

Our goal is a model that estimates ACs. This decomposes into finding matching points and predicting their local affine matrices. We obtain PCs from the \textit{frozen} LoMa matcher. To obtain a local affine matrix for each detected keypoint, we append an affine head (\cref{subsec:architecture}) to LoMa. In particular, we predict $\hat{\mat{A}}_i^{\mathcal A},\hat{\mat{A}}_j^{\mathcal B}\in\mathbb R^{2\times2}$ for $\mat{x}_i^{\mathcal A}$ and $\mat{x}_j^{\mathcal B}$, respectively. Each can be read as a local frame mapping a shared canonical patch onto the neighborhood of its keypoint, so that composing them maps the neighborhood of $\mat{x}_i^{\mathcal A}$ onto that of $\mat{x}_j^{\mathcal B}$.

Given a PC $(i,j)$, the affine transformation of the AC is obtained by composing the two per-keypoint matrices,
\begin{equation}
\hat{\mat{A}}_{ij}
=
\hat{\mat{A}}^{\mathcal B}_j
\left(
\hat{\mat{A}}^{\mathcal A}_i
\right)^{-1},
\label{eq:relative_affine}
\end{equation}
which we compare to the GT $\mat{A}_{ij}$ of \cref{eq:jacobian} during training (see \cref{subsec:supervision}). While $\hat{\mat{A}}_{ij}$ could also be regressed directly per match, per-keypoint frames make the estimate inverse-consistent by construction, \ie~the reverse affine from $\mathcal B$ to $\mathcal A$ is exactly $\hat{\mat{A}}_{ij}^{-1}$, as the model is permutation equivariant.

Note that \cref{eq:relative_affine} only constrains the \emph{composition} of the two per-keypoint predictions: for any invertible $\mat{G}\in\mathbb R^{2\times2}$, replacing
$\hat{\mat{A}}_i^{\mathcal A}\!\to\!\hat{\mat{A}}_i^{\mathcal A}\mat{G}$ and $\hat{\mat{A}}_j^{\mathcal B}\!\to\!\hat{\mat{A}}_j^{\mathcal B}\mat{G}$ leaves $\hat{\mat{A}}_{ij}$, and hence the loss, unchanged. Training therefore never pins down $\hat{\mat{A}}_i^{\mathcal A}$ itself, only its relation to that of a matched partner. We study this gauge
freedom empirically in \cref{subsec:gauge-freedom}.

Each per-keypoint matrix must be parametrized to guarantee invertibility, as required by \cref{eq:relative_affine}. Following prior work~\cite{mishkin2018repeatability,sun2025learning}, we decompose $\mat{A}\in\mathbb{R}^{2\times2}$ into a scale, an orientation, and a shape,
\begin{equation}
\setlength{\arraycolsep}{3pt} %
\mat{A}
= \lambda\, \mat{R}(\alpha)\,\mat{A}'
= \lambda
\begin{pmatrix}
\cos\alpha & -\sin\alpha \\
\sin\alpha & \cos\alpha
\end{pmatrix}
\begin{pmatrix}
e^{a} & b \\
0 & e^{-a}
\end{pmatrix},
\label{eq:decomp}
\end{equation}
where $\lambda>0$ is an isotropic scale (so that $\det\mat{A}=\lambda^2$), $\mat{R}(\alpha)$ a planar rotation, and $\mat{A}'$ a residual shape with $\det\mat{A}'=1$. We evaluate a simpler parametrization based on LU decomposition in \cref{subsec:ablations} and find that it performs almost as well, indicating that our approach is not highly sensitive to the parametrization.

\subsection{Affine Head Architecture}\label{subsec:architecture}
For every detected keypoint, the affine head takes the output descriptors of LoMa, and passes them through $K=3$ transformer layers of alternating self- and cross-attention. A per-keypoint MLP then regresses five numbers for the four DoF of \cref{eq:decomp}: scale (1), orientation (2, a direction vector normalized to unit length, which avoids the wraparound of regressing an angle directly), and shape (2, the stretch $a$ and the shear $b$), to compose $\hat{\mat{A}}$. To keep $\hat{\mat{A}}$, and hence its inverse in \cref{eq:relative_affine}, well conditioned, the scale $\lambda=e^{\Delta\tanh(\cdot)}$ and the stretch $a=\Delta\tanh(\cdot)$ are bounded with fixed $\Delta=2$, while the shear $b$ is regressed directly. The final layer of the prediction head is zero-initialized, so every keypoint starts with an identity affine frame. Because the head attends across views, the per-keypoint affine frames are conditioned on the partner image. We analyze this in \cref{subsec:gauge-freedom} and propose in the Suppl. (\cref{supp:transport}) a method to obtain stable per-view affine frames, useful for \eg~Structure-from-Motion (SfM).

\subsection{Supervision}\label{subsec:supervision}

For each training pair we have a GT projective ($z$-axis) depth map $D^{\mathcal A}$, camera intrinsics $\mat{K}^{\mathcal A},\mat{K}^{\mathcal B}$, and relative pose $(\mat{R},\mat{t})$, which make the warp $\phi$ of \cref{eq:jacobian} explicit,
\begin{equation}
    \phi(\mat{x}) = \pi\Big(\mat{K}^{\mathcal B}\big(\mat{R}\,D^{\mathcal A}(\mat{x})\,(\mat{K}^{\mathcal A})^{-1}\tilde{\mat{x}} + \mat{t}\big)\Big),
    \label{eq:depth_warp}
\end{equation}
where $\tilde{\mat{x}}$ is $\mat{x}$ in homogeneous coordinates and $\pi$ denotes the perspective division. From $\phi$ we derive the set of GT matches $\mathcal{M}$ and affine transformations $\mat{A}$. Rather than estimating a surface normal and differentiating the homography it induces, we differentiate the warp itself, estimating $\mat{A}_{ij}$ by sampling $\phi$ on a pixel grid of step $h$ and taking forward differences,
\begin{equation}
\begin{split}
    \mat{A}_{ij} \approx \frac1h\Big[&\,\phi\big(\mat{x}_i^{\mathcal A}+h\,\mat{e}_1\big)-\phi\big(\mat{x}_i^{\mathcal A}\big) \;\Big| \\
    &\;\phi\big(\mat{x}_i^{\mathcal A}+h\,\mat{e}_2\big)-\phi\big(\mat{x}_i^{\mathcal A}\big)\,\Big],
\end{split}
    \label{eq:findiff}
\end{equation}
with $\mat{e}_1,\mat{e}_2$ the standard basis of $\mathbb R^2$ and columns indexing the partial derivatives. We use $h=1$\,px and express the result in normalized image coordinates (see Suppl., \cref{supp:implementation}).

The GT matches $\mathcal M$ are the mutual nearest neighbors of the depth-induced warp $\phi$, restricted to matches whose local Jacobian is well defined, \ie, away from occlusion boundaries. We supervise the predicted AC of every $(i,j)\in\mathcal M$ directly against the GT affine $\mat{A}_{ij}$,
\begin{equation}\label{eq:loss}
    \mathcal{L}(\mat{A}, \hat{\mat{A}})=\frac{1}{|\mathcal M|}\sum_{(i,j)\in\mathcal M}\rho\!\left(\mat{A}_{ij},\hat{\mat{A}}_{ij}\right).
\end{equation}

For $\rho$ we use the robust Charbonnier loss, a special case of the general robust loss of \citet{barron2019general},
\begin{equation}
    \rho(\mat{A}_{ij}, \hat{\mat{A}}_{ij}) = \sqrt{\lVert\mat{A}_{ij}-\hat{\mat{A}}_{ij}\rVert_F^2+\delta^2} - \delta,
\end{equation}
with smoothing constant $\delta=10^{-3}$. We show in \cref{subsec:ablations} that the plain squared Frobenius loss $\rho(\mat{A}_{ij}, \hat{\mat{A}}_{ij}) =\lVert\mat{A}_{ij}-\hat{\mat{A}}_{ij}\rVert_F^2$ leads to worse results. %

\section{Results}\label{sec:results}

We compare to state-of-the-art techniques for AC estimation on a wide range of benchmarks. We evaluate the accuracy of the ACs directly (\cref{subsec:results:affine-acc}), and then their utility inside minimal solvers for relative pose (\cref{subsec:results:relpose}), visual localization (\cref{subsec:results:visloc}), and recovering surface normals (\cref{subsec:results:normal-est}). Throughout this section, the \colorbox{firstcolor}{best}, \colorbox{secondcolor}{second}, and \colorbox{thirdcolor}{third} ranked results are highlighted. Additional details are given in the Suppl. (\cref{supp:evaluation-details}).

\paragraph{Implementation details.} We train \ours~for 100K steps using the AdamW~\cite{adamw:2019} optimizer with a fixed learning rate of $2\times10^{-4}$ and a global batch size of 64, similar to LoMa. Training on $4\times$H200 GPUs at a resolution of $560\times560$ takes roughly a day. Varying the LoMa backbone, we release three models, \ours-$\{\mathrm{B},\mathrm{L},\mathrm{G}\}$, of increasing embedding dimension. The training data, detailed in the Suppl. (\cref{supp:training-data}), is a large mix of 3D datasets with GT depth, camera intrinsics, and poses. 

\paragraph{Baselines.} We compare to estimators of \emph{sparse} ACs: 

\begin{itemize}
    \item \textbf{\baffnet:} KeyNet~\cite{barroso2019key}  keypoints, HardNet~\cite{hardnet:2017} descriptors, and affine shape from AffNet~\cite{mishkin2018repeatability}.
    \item \textbf{\bsesti:} SIFT~\cite{lowe2004distinctive} keypoints, HardNet descriptors, and scale and orientation from S3Esti~\cite{yan2022learning}, assuming identity shape, \ie~$\mat{A}=\lambda\,\mat{R}(\alpha)$ with $\mat{A}'=\mat{I}$ in \cref{eq:decomp}.
    
    \item \textbf{\sunetalcite:}\footnote{We use the public \href{https://github.com/stilcrad/DenseAffine}{code} with bug fixes (see \cref{supp:sun-repro}).} Cropping patches around sparse DKM~\cite{edstedt2023dkm} PCs and estimating one affine frame per patch.

    \item \textbf{\blomaff:} LoMa PCs with scale and orientation from S3Esti and shape from AffNet.    
\end{itemize}
Sparse ACs can also be obtained by differentiating a dense matcher's warp $\phi$ directly (\cref{eq:findiff}). We include such ACs from RoMa v2~\cite{edstedt2026romav2} as a reference in \cref{tab:affine-frob,tab:normal}.

\subsection{Affine Accuracy}\label{subsec:results:affine-acc}

\begin{table}
\centering
\caption{\textbf{Accuracy of affine transformations.} We report median Frobenius ($\times100$) and rotational errors (lower is better).}
\label{tab:affine-frob}
\small

\setlength{\tabcolsep}{3pt}

\begin{tabular}{l@{\hspace{6pt}}rr rr}
\toprule
Method &
\multicolumn{2}{c}{HPatches} &
\multicolumn{2}{c}{Mega1500} \\
\cmidrule(lr){2-3}
\cmidrule(lr){4-5}
& Frobenius & Rot. & Frobenius & Rot. \\
\midrule

\multicolumn{5}{@{}l}{\refrow{\footnotesize\itshape Differentiated dense warp, at LoMa keypoints}} \\
\refrow{RoMa v2} & \refrow{2.82} & \refrow{0.46$^\circ$} & \refrow{5.04} & \refrow{0.75$^\circ$} \\
\midrule

\baffnet & 26.9 & 6.35$^\circ$ & 32.7 & 7.17$^\circ$ \\
\bsesti & 54.4 & 9.52$^\circ$ & 49.3 & 5.50$^\circ$ \\
\sunetal & 48.2 & 3.82$^\circ$ & 40.1 & 2.09$^\circ$ \\
\blomaff & 51.9 & 5.06$^\circ$ & 42.9 & 2.39$^\circ$ \\
\ours-B & \good 1.62 & \good 0.26$^\circ$ & \good 3.97 & \good 0.58$^\circ$ \\
\ours-L & \better 1.59 & \best 0.23$^\circ$ & \better 3.78 & \better 0.55$^\circ$ \\
\ours-G & \best 1.50 & \best 0.23$^\circ$ & \best 3.61 & \best 0.53$^\circ$ \\
\bottomrule
\end{tabular}
\end{table}

We begin by evaluating the accuracy of the predicted ACs directly, similar to \sunetalcite. \cref{fig:qualitative-patch} provides a qualitative comparison. GT affines are obtained differently on the two datasets. On HPatches~\cite{balntas2017hpatches} we use the \emph{analytic} Jacobian of the known homography, which involves neither depth nor the finite differences of \cref{subsec:supervision}, whereas on MegaDepth~\cite{li2018megadepth} we differentiate the depth-induced warp as during training. Following \sunetal, we restrict evaluation to matches whose predicted point already agrees with the GT warp (within 3 pixels at a canonical $480$-px scale) and whose GT affine is well defined (positive determinant), isolating the affine estimate itself. We additionally report scores that depend on the PCs in the Suppl. (\cref{subsec:ac-yield}). For every kept match, we compare $\hat{\mat{A}}_{ij}$ to the GT $\mat{A}_{ij}$ using the Frobenius error $\lVert\hat{\mat{A}}_{ij}-\mat{A}_{ij}\rVert_F$ and a residual rotation (\ie~the angle of the closest planar rotation to $\hat{\mat{A}}_{ij}\mat{A}_{ij}^{-1}$). We report the Frobenius error multiplied by 100 throughout, which is easier to read. \cref{tab:affine-frob} reports the median performance on HPatches and MegaDepth-1500 (\num{1500} outdoor pairs from the LoFTR split~\cite{sun2021loftr}). \ours~outperforms all sparse baselines by an order of magnitude in Frobenius error, and is also more accurate than ACs obtained by differentiating RoMa v2's dense warp (gray row).

\subsection{Relative Pose from Minimal AC Solvers}\label{subsec:results:relpose}

\begin{table}
\centering
\caption{\textbf{Essential matrix estimation from ACs.} We report area under the curve (AUC) of the pose error. For fair comparison against~\cite{sun2025learning} we use the 2AC solver inside GC-RANSAC~\cite{barath2021graph}.}
\label{tab:essential}

\setlength{\tabcolsep}{3pt}

\begin{tabularx}{3.0in}{Xrrrrrr}
\toprule
&
\multicolumn{3}{c}{Mega1500} &
\multicolumn{3}{c}{ScanNet1500} \\
\cmidrule(lr){2-4}
\cmidrule(lr){5-7}
Method &
\multicolumn{1}{c}{$5^\circ$} & 
\multicolumn{1}{c}{$10^\circ$} & 
\multicolumn{1}{c}{$20^\circ$} &
\multicolumn{1}{c}{$5^\circ$} & 
\multicolumn{1}{c}{$10^\circ$} & 
\multicolumn{1}{c}{$20^\circ$} \\
\midrule
\baffnet        & 46.4 & 58.7 & 67.5 & 6.8 & 13.2 & 20.4 \\
\bsesti & 46.1 & 58.3 & 67.2 & 5.7 & 11.4 & 18.4 \\
\sunetalname$^\dagger$ & 61.5 & 73.1 & 81.1 & 16.4 & 30.7 & 44.7 \\
\refrow{\sunetalname$^\ddagger$} & \refrow{65.5} & \refrow{77.9} & \refrow{86.3} & \refrow{---} & \refrow{---} & \refrow{---} \\
\blomaff & 63.4 & 77.2 & 86.7 & 28.1 & 50.0 & 67.9 \\
\ours-B       & \good 64.8 & \good 78.1 & \good 87.2 & \good 28.4 & \good 50.4 & \good 68.3 \\
\ours-L       & \best 66.0 & \best 78.9 & \best 87.8 & \better 29.7 & \better 51.9 & \better 69.8 \\
\ours-G       & \better 65.7 & \better 78.5 & \better 87.5 & \best 30.0 & \best 52.3 & \best 70.2 \\
\bottomrule
\multicolumn{7}{l}{\footnotesize $^\dagger$Results reproduced by the \href{https://github.com/stilcrad/DenseAffine}{official code}.} \\
\multicolumn{7}{l}{\footnotesize $^\ddagger$Officially reported numbers from the paper~\cite{sun2025learning}.} \\
\end{tabularx}
\end{table}

We now use the predicted ACs inside minimal solvers for relative pose, under three increasingly general camera models. Each setting substitutes ACs for the PCs a standard solver would otherwise consume, shrinking the minimal sample to 2, 7, and 9 ACs, respectively.

\paragraph{Calibrated two-view (2AC).} We consider relative pose estimation between two calibrated cameras through the estimation of the essential matrix~\cite{Hartley_Zisserman_2004}. Following \sunetal, we use GC-RANSAC~\cite{barath2021graph} with the 2AC solver~\cite{Barth2018EfficientRO}. We evaluate on MegaDepth-1500 and ScanNet-1500~\cite{dai2017scannet} (\num{1500} indoor pairs from~\citet{sarlin2020superglue}). \ours~outperforms existing AC estimators, particularly on ScanNet, where it exceeds \sunetal, who only provide outdoor weights, by 25.5 AUC@$20^\circ$. Most of this gap comes from LoMa's PCs, as \blomaff~shows. On the same matches, our affine head gives a consistent but smaller gain.

\paragraph{Rolling shutter (7AC).} Rolling-shutter cameras break the single-center projection model and require their own solvers, for both absolute~\cite{saurer2015minimal,albl2015r6p,albl2016rolling,albl2020rolling} and relative~\cite{dai2016rollingshuttercamerarelative,hruby2025singlescanline} pose. Following \citet{Barath2026RollingShutter}, we evaluate rolling shutter relative pose estimation accuracy on TUM-RS~\cite{Schubert_2019} using stride 20, 5K iterations of SupeRANSAC~\cite{barath2025superansacransacrule} with MSAC scoring and PROSAC sampling~\cite{chum2005matching}, and \num{2048} ACs. As shown in \cref{tab:rollingshutter}, moving from 20 PCs to 7 ACs improves accuracy for every estimator, and \ours-B gives the best result overall (71.6 AUC@$20^\circ$). \sunetal remain competitive here, driven by accuracy of DKM PCs.  

\begin{table}[t]
\centering
\caption{\textbf{Rolling shutter pose estimation from PCs vs. ACs.} We report pose AUC (\%) at \ang{5}/\ang{10}/\ang{20} on TUM-RS~\cite{Schubert_2019} comparing the RS-20PC~\cite{dai2016rollingshuttercamerarelative} solver and the RS-7AC~\cite{Barath2026RollingShutter} solver.}
\label{tab:rollingshutter}
\begin{tabular}{ll rrr}
\toprule
Solver & Method & @\ang{5} & @\ang{10} & @\ang{20} \\
\midrule

\multirow{4}{*}{20PC}

& \bkeynet & 3.0 & 5.1 & 7.8 \\
& \bsift & 11.1 & 14.6 & 17.2 \\
& \bdkm & 41.2 & 47.5 & 52.0\\
& LoMa-B & 22.6 & 30.7 & 36.7 \\
\midrule
\multirow{5}{*}{7AC}
& \baffnet & 36.9 & 46.2 & 53.3 \\
& \bsesti & 37.6 & 44.4 & 49.6 \\
& \sunetal & \better 60.0 & \good 65.5 & \good 69.0 \\
& \blomaff & \good 58.4 & \better 66.3 & \better 70.9 \\
& \ours-B & \best 60.1 & \best 67.1 & \best 71.6\\
\bottomrule
\end{tabular}
\end{table}

\begin{table}[t]
\centering
\caption{\textbf{Generalized relative pose and scale from PCs vs. ACs.} We report AUC (\%) at 0.25\,m/2$^\circ$ on TUM RGB-D, with RANSAC either capped at 100 iterations (\emph{100 it.}) or uncapped (\emph{No cap}). \emph{Iters.}\ is the mean number of RANSAC iterations when uncapped.}
\label{tab:generalized-relpose}
\begin{tabular}{ll rrr}
\toprule
& & & \multicolumn{2}{c}{AUC@0.25\,m, $2^\circ$} \\
\cmidrule(lr){4-5}
Solver & Method &
\multicolumn{1}{c}{Iters.} &
\multicolumn{1}{c}{No cap} &
\multicolumn{1}{c}{100 it.} \\
\midrule

\multirow{4}{*}{26PC}

& \bkeynet & 930 & 75.7 & 61.2 \\
& \bsift & 911 & \good 76.2 & 61.2 \\
& \bdkm & 303 & 73.3 & \better 72.1 \\
& LoMa-B & 1009 & \better 78.7 & 64.7 \\
\midrule
\multirow{5}{*}{9AC}
& \baffnet & 2288 & 56.3 & 19.3 \\
& \bsesti & 329 & 70.6 & 53.3 \\
& \sunetal & \better 263 & 70.1 & \good 64.8 \\
& \blomaff & \good 292 & 72.3 & 57.1 \\
& \ours-B & \best 73 & \best 79.4 & \best 76.7 \\
\bottomrule
\end{tabular}
\end{table}

\paragraph{Multi-camera rigs (9AC).} Following \citet{xu2025generalized},\footnote{Reimplemented from the paper. No code available as of this writing.} we consider the problem of registering two view-graphs of four posed frames each with a generalized relative-pose-and-scale solver. We evaluate on TUM RGB-D~\cite{sturm2012benchmark} and compare the AUC of a 26PC solver against a 9AC solver in \cref{tab:generalized-relpose}. For a given matcher, both solvers receive identical matches, while the AC solver additionally receives the predicted affine of each match. \ours-B~improves accuracy over the point baseline while requiring approximately $14\times$ fewer RANSAC iterations. AffNet, in contrast, needs $2.5\times$ \emph{more} iterations with 9 ACs than with 26 PCs.

\subsection{Visual Localization}\label{subsec:results:visloc}

We evaluate our affine estimates inside the absolute-pose pipeline of \citet{valtonen-ornhag-jaenal-2026-eccv}, substituting only the predicted local frames and leaving the reconstruction, detector (SuperPoint~\cite{detone2018superpoint}), matcher (LightGlue~\cite{lindenberger2023lightglue}), retrieval, and RANSAC configuration untouched. We compare three minimal solvers: \solvsift~\cite{Ventura_2024_CVPR}, using scale and orientation and assuming a known focal length, \solvac~\cite{valtonen-ornhag-jaenal-2026-eccv}, using the full affine frame and also recovering the focal length, and \solvpac~\cite{Ventura_2023_ICCV}, using the full affine frame with a known focal length. We report results on Cambridge Landmarks~\cite{kendall2015posenet} in \cref{tab:cambridge_abspose_short}. We disable local optimization (LO). With LO enabled, the gains for \solvac and \solvpac become negligible beyond about 100 iterations and remain at small budgets, while those for \solvsift persist (Suppl., \cref{tab:cambridge_abspose_lo,fig:budget-curve}). Replacing S3Esti's estimates with ours slightly improves accuracy for all solvers.

\begin{table}[t]
\centering
\small
\caption{\textbf{Absolute pose on Cambridge Landmarks.} %
We report the mean over the five scenes, with 95\% confidence intervals over 100 RANSAC seeds.}
\begin{tabular}{llcc}
\toprule
Solver & Method & 25\,cm, $2^\circ$ & 50\,cm, $5^\circ$ \\
\midrule
\refrow{P4Pf} & \refrow{Points only} & \refrow{41.4\,\tiny$\pm$0.07} & \refrow{63.3\,\tiny$\pm$0.05} \\
\midrule
\multirow{2}{*}{\solvsift}
 & S3Esti & 44.8\,\tiny$\pm$0.17 & 59.6\,\tiny$\pm$0.16 \\
 & \ours-B & \bfseries 49.9\,\tiny$\pm$0.15 & \bfseries 65.5\,\tiny$\pm$0.14 \\
\midrule
\multirow{2}{*}{\solvac}
 & S3Esti+AffNet & 39.5\,\tiny$\pm$0.15 & 61.5\,\tiny$\pm$0.13 \\
 & \ours-B & \bfseries 40.6\,\tiny$\pm$0.14 & \bfseries 62.1\,\tiny$\pm$0.12 \\
\midrule
\multirow{3}{*}{\solvpac}
 & S3Esti & 67.0\,\tiny$\pm$0.12 & 84.7\,\tiny$\pm$0.09 \\
 & S3Esti+AffNet & 67.1\,\tiny$\pm$0.12 & 84.6\,\tiny$\pm$0.11 \\
 & \ours-B & \bfseries 67.4\,\tiny$\pm$0.12 & \bfseries 84.8\,\tiny$\pm$0.09 \\
\bottomrule
\end{tabular}
\label{tab:cambridge_abspose_short}
\end{table}

\subsection{Normal Estimation}\label{subsec:results:normal-est}

Next, we evaluate the accuracy of our ACs through their utility for surface normal estimation. We recover the surface normal from a single AC and the known relative pose based on the optimal solver of \citet{hajder2023fast}. We keep only correspondences consistent with the known relative pose (Sampson distance $<5$ px), decoupling normal accuracy from matching precision. We compare the recovered normal to a reference normal obtained by fitting a plane to the GT depth in a small window around the keypoint, using the sign-invariant angle between the two unit normals as the error. \cref{tab:normal} reports the area under this per-correspondence angular-error curve (AUC). We find that \ours~outperforms the dedicated AC estimators, which fail on this benchmark. \ours-L and \ours-G also outperform normals recovered from RoMa v2's dense warp.

\begin{table}
\centering
\caption{\textbf{Normal estimation from ACs.} We report area under the curve (AUC), higher is better. }
\label{tab:normal}

\setlength{\tabcolsep}{3pt}

\begin{tabular}{l@{\hspace{6pt}}rrr rrr}
\toprule
Method &
\multicolumn{3}{c}{Map-free~\cite{arnold2022map}} &
\multicolumn{3}{c}{Mega1500} \\
\cmidrule(lr){2-4}
\cmidrule(lr){5-7}
& 5$^\circ$ & 10$^\circ$ & 20$^\circ$
& 5$^\circ$ & 10$^\circ$ & 20$^\circ$ \\
\midrule

\multicolumn{7}{@{}l}{\refrow{\footnotesize\itshape Differentiated dense warp, at LoMa keypoints}} \\
\refrow{RoMa v2} & \refrow{5.9} & \refrow{14.1} & \refrow{27.8} & \refrow{4.3} & \refrow{12.7} & \refrow{29.2} \\
\midrule

\baffnet & 0.3 & 1.3 & 4.8 & 0.3 & 1.3 & 4.4 \\
\bsesti & 0.2 & 0.7 & 2.8 & 0.1 & 0.5 & 1.9 \\
\sunetal & 0.1 & 0.5 & 1.9 & 0.2 & 0.9 & 3.3 \\
\blomaff & 0.2 & 0.6 & 2.4 & 0.2 & 0.7 & 2.8 \\
\ours-B & \good 9.9 & \good 19.2 & \good 33.3 & \good 4.3 & \good 12.9 & \good 30.4 \\
\ours-L & \better 9.7 & \better 19.2 & \better 33.5 & \better 4.4 & \better 13.6 & \better 32.0 \\
\ours-G & \best 9.8 & \best 19.3 & \best 33.6 & \best 5.0 & \best 14.7 & \best 33.5 \\
\bottomrule
\end{tabular}
\end{table}

\begin{figure*}[t]
\centering
\includegraphics[width=0.99\linewidth]{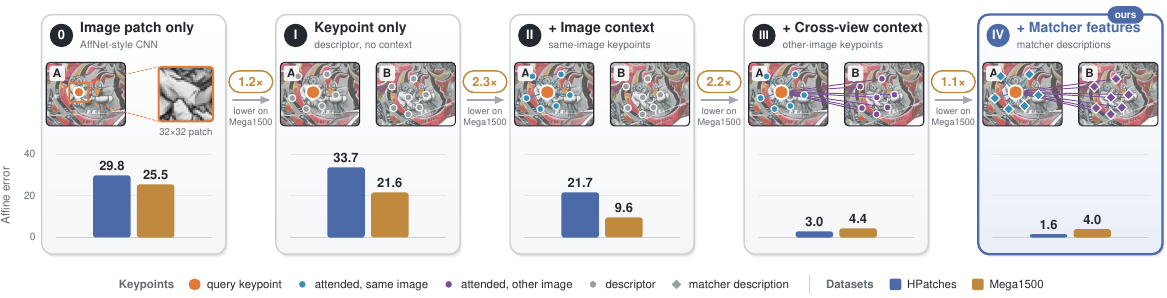}
\caption{\textbf{Affine estimation benefits from global context.} A CNN on the raw image patch around each keypoint (0), trained with our loss and setup in the style of AffNet, is the weakest. Training a head on the keypoint descriptions (I) is only slightly better and comparable to AffNet. As the attention context is extended, first to the same image (II) and then to the other image (III), the error on MegaDepth-1500 is roughly halved at each step. Reading the matcher's features instead of the descriptor's (IV) reduces it further.}
\label{fig:model-context}
\end{figure*}

\section{Analysis}

In this section, we study our design choices (\cref{subsec:ablations}) and the gauge freedom induced by our supervision (\cref{subsec:gauge-freedom}).

\subsection{Design Choices}\label{subsec:ablations}

\paragraph{Local versus global information.} To our knowledge, all prior dedicated AC estimators use local patches. In \cref{fig:model-context}, we increase the context from local image patches (0), to local keypoint descriptors, to global (II), to cross-view (III), each step of increased context roughly halving the affine error on MegaDepth-1500. Moving from the descriptor features to the matcher (IV) reduces it slightly further. Although (I), (II), and (III) have the same input descriptors, the additional attention context substantially decreases the error. Interestingly, both (0), using local patch information, and (I), using isolated keypoint descriptors, manage to achieve AffNet-level performance through direct supervision and jointly predicting scale, shape, and orientation.

\paragraph{Parametrization.} AffNet and \sunetal both conclude that the parametrization of $\mat{A}$ is critical. While we find that using scale, shape, and orientation works well, in \cref{tab:ablation-affine}, we also try a simpler $\mat{L}\, \mat{U}$ decomposition,
\begin{equation}
\hat{\mat{A}}
= \mat{L}\, \mat{U}
= \begin{pmatrix}
1 & 0 \\
\gamma & 1
\end{pmatrix}
\begin{pmatrix}
e^{u_1} & \beta \\
0 & e^{u_2}
\end{pmatrix},
\label{eq:ludecomp}
\end{equation}
where $u_1, u_2 = \Delta\tanh(\cdot)$ with $\gamma,\beta$ regressed directly (unbounded). It can only represent frames whose top-left entry is positive, so, for example, it cannot express a single frame rotated by more than $90^\circ$. \cref{eq:decomp} has no such limit. Nonetheless, it performs comparably despite requiring no trigonometric parametrization, suggesting that our formulation is less sensitive to parametrization than prior work.

\begin{table}[t]
\centering
\caption{\textbf{Supervision and parametrization.} Median affine error (Frobenius, $\times100$) on MegaDepth-1500 for each combination of loss $\rho$ (\cref{eq:loss}) and parametrization of the per-keypoint frames, with $\Delta=\mat{A}_{ij}-\hat{\mat{A}}_{ij}$. Our default is shaded.}
\label{tab:ablation-affine}
\small
\begin{tabular}{l cc}
\toprule
\diagbox[height=1.6\line, innerleftsep=0pt]{Parametrization}{Loss $\rho$} & $\lVert\Delta\rVert_F^2$ & $\sqrt{\lVert\Delta\rVert_F^2+\delta^2}-\delta$ \\
\midrule
$\mat{L}\,\mat{U}$, \cref{eq:ludecomp} & 11.9 & 4.3 \\
$\lambda\,\mat{R}(\alpha)\,\mat{A}'$, \cref{eq:decomp} & 11.2 & \cellcolor{gray!25}\bfseries 4.0 \\
\bottomrule
\end{tabular}
\end{table}

\paragraph{Supervision.} In \cref{tab:ablation-affine}, we compare using the squared Frobenius loss $\lVert\cdot\rVert_F^2$ as $\rho$ in \cref{eq:loss} to using the robust Charbonnier loss of \cref{subsec:supervision}, which for our small smoothing constant $\delta$ is essentially the unsquared Frobenius norm. We find that the latter more than halves the affine error.

\paragraph{Architecture.} In \cref{tab:ablation}, we study the effect of changing the architecture. We begin by jointly training matching and affine estimation from scratch (A), which leads to strong results. Initializing from LoMa and then fine-tuning with matching and affine estimation supervision jointly (B) does not help. While freezing the matcher (C) slightly improves matching accuracy, the linear head is insufficient. However, increasing the capacity of the affine head to a shallow transformer (E) leads to both strong matching and strong affine estimation performance. D slightly outperforms E on affine error but requires training the backbone, slightly deteriorating matching performance. We adopt E, which is ${\sim}4\times$ cheaper to train and leaves the matcher untouched.

\begin{table}[t]
\centering
\caption{\textbf{Architecture.} Affine error (Frobenius, $\times100$) and relative pose AUC@$20^\circ$ (5 PCs) on MegaDepth-1500. \snow~and \flame~denote a frozen and a trained matcher. A trained matcher also receives LoMa's matching loss. Our default is shaded.}
\label{tab:ablation}
\small
\setlength{\tabcolsep}{4pt}
\begin{tabular}{c l c l rr}
\toprule
& Initialization & Matcher & Affine head & Frob.\,$\downarrow$ & AUC\,$\uparrow$ \\
\midrule
A & Scratch & \flame & Linear      & 4.6  & 83.2 \\
B & LoMa    & \flame & Linear      & 4.8  & 83.2 \\
C & LoMa    & \snow  & Linear      & 12.2 & \bfseries 83.6 \\
\midrule
D & Scratch & \flame & Transformer & \bfseries 3.8 & 83.2 \\
\rowcolor{gray!25}
E & LoMa    & \snow  & Transformer & 4.0  & \bfseries 83.6 \\
\bottomrule
\end{tabular}
\end{table}

\paragraph{Orientation.} Estimating the full affine frame includes orientation, which prior local methods
predict with a separate network. Training and evaluating on independently rotated pairs, we find that a
frozen rotation-invariant matcher (LoMa-R~\cite{nordstrom2026whohandlesorientation}) cannot support
orientation estimation, since its features discard exactly the quantity to be predicted, whereas unfreezing
it recovers accurate orientation and good matches under rotation (see \cref{supp:rotation}).

\subsection{Gauge Freedom} \label{subsec:gauge-freedom}

Our approach supervises the ACs $\hat{\mat{A}}_{ij}$ directly, never the keypoint local frames $\hat{\mat{A}}_i^{\mathcal A}$.
As noted in \cref{subsec:pred-acs}, this leaves a ``gauge freedom'' in the local predictions: $\hat{\mat{A}}_{ij}=(\hat{\mat{A}}_j^\mathcal{B}\mat{G})(\hat{\mat{A}}_i^{\mathcal{A}}\mat{G})^{-1}$ for any invertible $\mat{G}$, which may depend on both keypoints and both images.
To illustrate the gauge freedom, we show in the Supplementary (\cref{fig:gauge}(b)) how the prediction of $\hat{\mat{A}}_i^\mathcal{A}$ varies as we change the partner image $\mat{I}^\mathcal{B}$.
We quantify this by the residual gauge $\mat{G}_{pq}=(\hat{\mat{A}}_p)^{-1}\hat{\mat{A}}_q$ between the frames predicted for the same keypoint under two different partner images $p$ and $q$, which is the identity if and only if the two agree. As a rough baseline, we consider that even for a fully resolved gauge the estimates would differ by $\sqrt{2}\,\epsilon$ to $2\epsilon$ for estimation error $\epsilon$ (to first order, using the error of \cref{tab:affine-frob} as a proxy). The measured median on the HPatches sequences we evaluate is around ten times larger, indicating that gauge freedom is unresolved.

\section{Limitations and Future Work}
\begin{itemize}
    \item While the additional context from the image pair improves performance, it couples the local affine frame to the paired image. Although stable per-view affine frames can be obtained when desired, such as in SfM, by averaging the estimates across pairs (\cref{supp:transport}), this affects performance. Studying this and the gauge freedom from our supervision constitutes interesting future work.
    
    \item We use LoMa's PCs rather than those of RoMa v2, which are often more accurate. Although this yields substantial runtime improvements (\cref{subsec:runtime}), leveraging dense matchers and feed-forward reconstruction models~\cite{wang2025vggt} for AC estimation constitutes interesting future work.
    
    \item With LO and large iteration budgets, the gains from accurate ACs are modest (\cref{subsec:results:visloc}). We find ACs to be most beneficial in solvers requiring a large number of PCs, such as rolling-shutter relative pose (\cref{tab:rollingshutter}). Identifying problems where ACs are vastly superior to PCs constitutes interesting future work.
\end{itemize}

\section{Conclusions}

We revisit AC estimation and go beyond local patches. Our formulation allows a simpler pipeline that directly regresses the full 4-DoF affine matrix. Systematic ablations show that global context is crucial for affine estimation. Our method, \ours, adds a lightweight affine head to LoMa's matcher, achieving an order of magnitude lower affine error than AffNet and \sunetal, with consistent downstream gains.

\section*{LLM Usage Statement}

\paragraph{We used Claude by Anthropic through Claude Code  for:} 
\begin{itemize}
    \item Generating code for the experiments in \cref{sec:results}.
    \begin{itemize}
        \item Source code was read by the authors and evaluations were based on previous work.
        \item The core implementation of the model and loss was done without the assistance of LLMs.
    \end{itemize}
    \item Checking for typos.
    \item Assist with editing the manuscript for clarity and writing quality. 
    \item Identifying relevant literature (checked by the authors).
    \item Generating the TikZ and Matplotlib code for the figures.
\end{itemize}

\paragraph{We did not use LLMs for:}
\begin{itemize}
    \item Helping develop theoretical models or conceptual frameworks.
    \item Suggesting presentation structure or style.
    \item Designing or providing feedback on research methodology or experiments.
    \item Proposing or refining hypotheses.
    \item Assisting in natural-language translation.
    \item Interpreting results.
    \item Drafting any part of the paper. The authors wrote all sections of the paper and only then received assistance from the LLM to refine it.
    \item Formatting references.
\end{itemize}

\paragraph{We have reviewed all LLM-assisted work.} We take responsibility for the final content of this work, including text, claims or artifacts produced with the aid of generative AI.

\section*{Acknowledgements}
This work was supported by the Wallenberg Artificial
Intelligence, Autonomous Systems and Software Program
(WASP), funded by the Knut and Alice Wallenberg Foundation, by the strategic research environment ELLIIT, funded by the Swedish government, and by the National Science Foundation under Award No. 2144822. 
The computational resources were provided by the
National Academic Infrastructure for Supercomputing in
Sweden (NAISS) at C3SE, partially funded by the Swedish Research
Council through grant agreement no.~2022-06725, and by
the Berzelius resource, provided by the Knut and Alice Wallenberg Foundation at the National Supercomputer Centre.

{
    \small
    \bibliographystyle{ieeenat_fullname}
    \bibliography{main}
}
\appendix
\clearpage
\setcounter{page}{1}
\maketitlesupplementary

\section{Implementation Details}\label{supp:implementation}

\paragraph{Architecture.} \ours~appends an affine head to a frozen LoMa~\cite{nordstrom2026lomalocalfeaturematching} matching pipeline: DaD keypoints~\cite{edstedt2025dad} (\num{2048} per image), the LoMa descriptor network, and a LightGlue-style~\cite{lindenberger2023lightglue} matcher of nine alternating self- and cross-attention layers with four heads. \ours-$\{\mathrm{B},\mathrm{L},\mathrm{G}\}$ differ only in the embedding dimension of the backbone (\num{256}, \num{512}, \num{1024}). The head applies $K=3$ further self- and cross-attention layers to the final matcher features, using the same keypoint positional encoding as the matcher in self-attention, followed by a per-keypoint MLP shared across both images (one hidden layer of the embedding width, GELU) that outputs five scalars parameterizing the four DoF of \cref{eq:decomp} (log-scale and shape exponent bounded by $\Delta\tanh(\cdot)$ with $\Delta=2$). The last layer is zero-initialized, so every keypoint starts at $\hat{\mat{A}}=\mat{I}$.

\paragraph{Supervision.} The GT affine of a match is the Jacobian of the depth-induced warp, computed by forward differences (\cref{eq:findiff}) on the dense warp at the $560\times560$ training resolution in normalized coordinates, \ie, $\mat{A}=\mat{S}_{\mathcal B}^{-1}\mat{J}\mat{S}_{\mathcal A}$ with $\mat{J}$ the pixel Jacobian and $\mat{S}=\mathrm{diag}(W/2,H/2)$. A Jacobian is kept only if the pixel and its forward neighbors are covisible (the projection agrees with the depth to within \num{5}\%), which rejects differences across occlusion boundaries, and rejected entries are masked out of the loss. Matches are the mutual nearest neighbors of the forward and backward warps at the keypoints. The loss of \cref{eq:loss} is applied to the composed relative affine $\hat{\mat{A}}_{ij}=\hat{\mat{A}}^{\mathcal B}_j(\hat{\mat{A}}^{\mathcal A}_i)^{-1}$ of every match. The per-keypoint frames are never supervised individually, which is the origin of the gauge freedom studied in \cref{subsec:gauge-freedom}.

\section{Training Data}\label{supp:training-data}

We train on the mix of \cref{tab:dataset_mix}. Each sample draws a dataset with probability proportional to the weight in \cref{tab:dataset_mix}, and then a pair by covisibility following standard matching sampling like in LoMa and RoMa. Images are resized to $560\times560$ and the intrinsics rescaled accordingly. None of the evaluation scenes are in the training data: MegaDepth scenes 0015 and 0022 are excluded from all MegaDepth-derived sets, the Map-free scenes used for normal estimation (\texttt{s00015}, \texttt{s00022}) are excluded from Map-free, and HPatches, ScanNet-1500, Cambridge Landmarks, TUM RGB-D and TUM-RS are not used for training.

\begin{table}[t]
\centering
\caption{\textbf{Training data}. We use a mix of indoor, outdoor and aerial training data, similar to \cite{wang2024dust3r,nordstrom2026lomalocalfeaturematching,edstedt2026romav2,wang2025vggt}.}
\label{tab:dataset_mix}
\setlength{\tabcolsep}{2pt}
\begin{tabular}{lcr}
\toprule
Datasets & Type / GT Source & Weight \\
\midrule
AerialMD~\cite{vuong2025aerialmegadepth} & Aerial / MVS & \num{10000} \\
ASE~\cite{AriaSynthEnv:2025} & Indoor / Graphics & \num{1000} \\
BlendedMVS~\cite{yao2020blendedmvs} & Aerial / Mesh & \num{10000} \\
FlyingThings3D~\cite{mayer2016large} & Synthetic / Graphics & \num{5000} \\
Hypersim~\cite{roberts2021hypersim} & Indoor / Graphics & \num{10000} \\
Map-free~\cite{arnold2022map} & Object-centric / MVS & \num{10000} \\
MegaDepth~\cite{li2018megadepth} & Outdoor / MVS & \num{10000} \\
MegaDepth-X~\cite{li2026longtail} & Outdoor / MVS & \num{10000} \\
MegaScenes~\cite{tung2024megascenes} & Outdoor / MVS & \num{10000} \\
MegaSynth~\cite{Jiang_2025_CVPR} & Indoor / Graphics & \num{1000} \\
MPSD~\cite{mpsd:2020} & Driving / MVS & \num{1000} \\
ScanNet++ v2~\cite{yeshwanth2023scannet++} & Indoor / Mesh & \num{10000} \\
SpatialVID~\cite{wang2025spatialvid} & Forward-motion / MVS & \num{100} \\
TartanAir V2~\cite{wang2020tartanair} & Synthetic / Graphics & \num{10000} \\
UnrealStereo4k~\cite{tosi2021unrealstereo4k} & Outdoor / Graphics & \num{100} \\
Virtual KITTI 2~\cite{gaidon2016virtual,cabon2020vkitti2} & Outdoor / Graphics & \num{100} \\
\bottomrule
\end{tabular}
\end{table}

\begin{figure*}[t]
\centering
\includegraphics[width=0.99\linewidth]{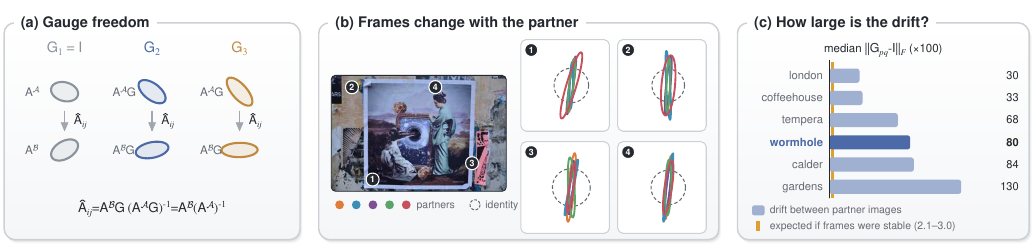}
\caption{\textbf{Gauge freedom.} \textbf{(a)}~Right-multiplying both frames of a match by a common $\mat{G}$ leaves $\hat{\mat{A}}_{ij}$ and the loss unchanged. \textbf{(b)}~Predicted frames of four keypoints of one image under five partner images (colors). The dashed circle is the identity. \textbf{(c)}~Median $\|\mat{G}_{pq}-\mat{I}\|_F$ ($\times100$) per HPatches sequence. The band is the range expected for stable frames.}
\label{fig:gauge}
\end{figure*}

\section{Evaluation Protocols}\label{supp:evaluation-details}

\paragraph{Reproducing \sunetalname\ }\label{supp:sun-repro}
We evaluate \sunetalcite with their public code and pretrained outdoor weights. The released demo had four implementation issues, which we fix in our code and that we will release publicly. First, the quality filter indexes the unsorted matches with positions from a sorted copy. Second, S3Esti's orientation, predicted in radians, is passed to a function expecting degrees. Third, S3Esti receives a raw grayscale crop instead of its three-scale normalized RGB pyramid. Fourth, the local affine frame (LAF) magnification factor is applied repeatedly. All other hyperparameters remain unchanged. For fairness, we also report the results reported by \sunetal~where available (\cref{tab:essential}).

\paragraph{Affine shapes.} Affines are compared in the normalized frame of each image, $\mat{A}=\mat{S}_{\mathcal B}^{-1}\mat{J}\mat{S}_{\mathcal A}$ with $\mat{S}=\mathrm{diag}(W/2,H/2)$ and $\mat{J}$ the pixel Jacobian. \cref{subsec:pixel-frame} repeats \cref{tab:affine-frob} in the pixel frame. The residual rotation and scale are read off $\hat{\mat{A}}_{ij}\mat{A}_{ij}^{-1}$ in the same frame. Every method is given a budget of \num{2048} keypoints and is evaluated on its own matches that pass the 3\,px test of \cref{subsec:results:affine-acc}. The baselines run at a long side of 1024\,px and our descriptor at $784\times784$. The HPatches GT is the analytic Jacobian of $H_{1\to i}$ at the matched pixel on the 56 viewpoint sequences (280 pairs, excluding \texttt{v\_artisans}, \texttt{v\_astronautis} and \texttt{v\_talent}, following LoFTR~\cite{sun2021loftr}). On MegaDepth-1500 it is the finite-difference Jacobian of the depth warp with the covisibility check used in training. Medians are taken over all kept matches pooled over the pairs of a dataset.

\begin{figure}[t]
\centering
\includegraphics[width=\linewidth]{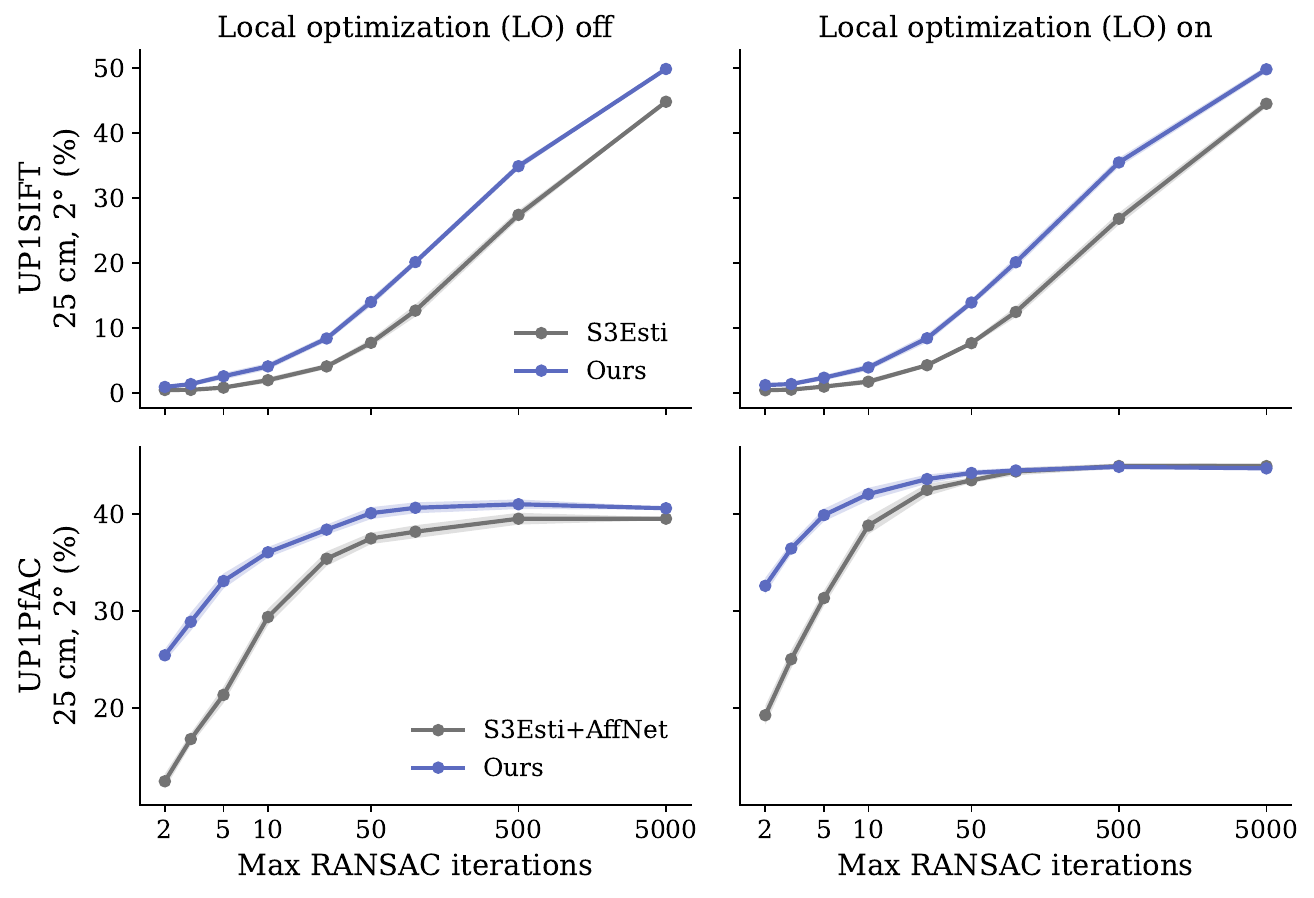}
\caption{\textbf{Accuracy against the RANSAC budget.} Cambridge Landmarks queries localized within 25\,cm, $2^\circ$ (mean over the five scenes, 95\% confidence bands) against the maximum number of GC-RANSAC iterations, with LO disabled (left) and enabled (right).}
\label{fig:budget-curve}
\end{figure}

\paragraph{Visual localization.} We use the implementation of~\cite{valtonen-ornhag-jaenal-2026-eccv} with GC-RANSAC~\cite{barath2021graph} on the re-triangulated COLMAP~\cite{schoenberger2016sfm} models of Cambridge Landmarks (five scenes, \num{1024}\,px). Each query is localized from its 10 NetVLAD~\cite{arandjelovic2016netvlad} references in hloc~\cite{sarlin2019coarse} with SuperPoint (\num{4096} keypoints) and LightGlue matches. Reconstruction, matches, retrieval and RANSAC are identical in all rows, and only the source of the frames differs. For \ours-B we run the model on every (query, reference) pair at the SuperPoint keypoints, discard LoMa's own matches, and read out the predicted frames at the LightGlue matches. Our frames depend on the pair, so they are computed per (query, reference) pair instead of once per image. \solvsift uses only the scale and orientation of each frame, read out from the same forward pass of the pair, thus requiring only S3Esti, while \solvac and \solvpac use the relative affine $\hat{\mat{A}}_{ij}$ of each match (\cref{eq:relative_affine}). For these two, the baseline combines S3Esti's scale and orientation with AffNet's shape. We report, for each RANSAC seed, the mean over scenes of the per-scene accuracy (scaled by the fraction of localized queries), with 95\% confidence intervals over 100 seeds, without LO in \cref{tab:cambridge_abspose_short} and with the default configuration in \cref{tab:cambridge_abspose_lo}. The points-only baseline uses the same matches with the P4Pf solver. \cref{fig:budget-curve} varies the RANSAC budget: with \solvsift our frames are more accurate at every budget, whereas with \solvac they help mostly at small budgets (\num{32.6}\% against \num{19.3}\% at 2 iterations with LO), and with LO enabled the two coincide from about 100 iterations on, so at the default budget the source of the frames does not matter for \solvac.

\begin{table}[t]
\centering
\small
\caption{\textbf{Absolute pose on Cambridge Landmarks with the default GC-RANSAC configuration (LO enabled).} Same protocol as \cref{tab:cambridge_abspose_short}.}
\begin{tabular}{llcc}
\toprule
Solver & Method & 25\,cm, $2^\circ$ & 50\,cm, $5^\circ$ \\
 \midrule
\refrow{P4Pf} & \refrow{Points only} & \refrow{41.4\,\tiny$\pm$0.04} & \refrow{63.7\,\tiny$\pm$0.02} \\
\midrule
\multirow{2}{*}{\solvsift}
 & S3Esti & 45.0\,\tiny$\pm$0.15 & 59.7\,\tiny$\pm$0.15 \\
 & \ours-B & \bfseries 49.8\,\tiny$\pm$0.15 & \bfseries 65.5\,\tiny$\pm$0.16 \\
\midrule
\multirow{2}{*}{\solvac}
 & S3Esti+AffNet & 44.7\,\tiny$\pm$0.07 & 66.0\,\tiny$\pm$0.06 \\
 & \ours-B & 44.7\,\tiny$\pm$0.06 & 66.1\,\tiny$\pm$0.07 \\
\midrule
\multirow{3}{*}{\solvpac}
 & S3Esti & 67.1\,\tiny$\pm$0.13 & 84.7\,\tiny$\pm$0.10 \\
 & S3Esti+AffNet & 66.8\,\tiny$\pm$0.13 & 84.7\,\tiny$\pm$0.09 \\
 & \ours-B & \bfseries 67.2\,\tiny$\pm$0.13 & \bfseries 84.8\,\tiny$\pm$0.09 \\
\bottomrule
\end{tabular}
\label{tab:cambridge_abspose_lo}
\end{table}

\paragraph{Generalized relative pose and scale.} We follow the view-graph registration protocol of~\cite{xu2025generalized} on TUM RGB-D~\cite{sturm2012benchmark} (\texttt{freiburg2\_xyz}/\texttt{rpy}), with our own implementation of the solvers as the authors' code is not available. A pair consists of two view-graphs of four posed frames each, the second in a random similarity frame (rotation, translation in $[-1,1]$\,m, scale in $[0.2,5]$) that is to be recovered. PC and AC solvers receive identical matches, the latter also the predicted affine of each match. RANSAC samples 26 PCs or 9 ACs, scores hypotheses with the Sampson error (2\,px) and refines the inliers with a null-space solver. AUC is the area under the recall curve of $\max(\epsilon_R/2^\circ,\epsilon_t/0.25\mathrm{m})$, pooled over both sequences.

\paragraph{Surface normals.} We keep correspondences whose Sampson distance to the known relative pose is below \num{5}\,px, which decouples normal accuracy from matching precision. The $\mathcal A$-side keypoint is backprojected with the GT depth, and the predicted affine (in calibrated coordinates) is passed with the GT pose to the optimal solver of \citet{hajder2023fast}. The reference normal is a plane fit (PCA) to the 3D points of a $9\times9$ window around the keypoint, using pixels within \num{5}\% of the center depth. On Map-free we use the validation scenes, which are excluded from training. We report the area under the curve of the sign-invariant angle between the two unit normals.

\section{Additional Experiments}

\subsection{Runtime}\label{subsec:runtime}

In \cref{tab:runtime}, we compare the matching speed of \ours~to \sunetal~and RoMa v2. As matching dominates the wall-clock time in SfM and visual localization, we only consider it.\footnote{See for example \href{https://github.com/zju3dv/EfficientLoFTR/issues/1}{this discussion} by P.E. Sarlin regarding ELoFTR.} Depending on model size, \ours~is roughly $9$--$13\times$ faster than \sunetal~and $8$--$12\times$ faster than RoMa v2. 

\begin{table}
\centering
\caption{\textbf{Runtime.} AC estimation speed on an H200 running a single image pair per batch (higher is better).}
\label{tab:runtime}
\small

\begin{tabular}{l r}
\toprule
Method
& Throughput (pairs / second) \\
\midrule

\sunetal & 3.9 \\
RoMa v2 & 4.3 \\
\ours-B & \best 51.1 \\
\ours-L & \better 40.5 \\
\ours-G & \good 35.7 \\

\bottomrule
\end{tabular}
\end{table}

\subsection{End-to-End AC Yield}\label{subsec:ac-yield}

The protocol of \cref{subsec:results:affine-acc} keeps only matches whose point is within \num{3}\,px of the GT warp, so the errors of \cref{tab:affine-frob} are conditional on a correct point match and say nothing about how often a method delivers a usable AC. We therefore also report an unconditional measure.

\paragraph{Protocol.} Each method gets a budget of $N=\num{2048}$ keypoints per image and produces at most $N$ ACs. An AC $(\mat{x}_i,\mat{x}_j,\hat{\mat{A}}_{ij})$ is \emph{$(\tau_p,\tau_A)$-correct} if its point and its affine are both right,
\begin{equation}
    \lVert\mat{x}_j-\phi(\mat{x}_i)\rVert < \tau_p
    \quad\text{and}\quad
    \lVert\hat{\mat{A}}_{ij}-\mat{A}_{ij}\rVert_F < \tau_A ,
    \label{eq:ac-correct}
\end{equation}
with $\phi$ the GT warp. The \emph{AC yield} is the fraction of the budget $N$ that is correct, so a method cannot gain by matching conservatively. Setting $\tau_A=\infty$ gives the ordinary point recall (\emph{Point} column). We fix $\tau_p=\num{3}\,$px and use $\tau_A=0.25$ and $0.1$ (roughly $10^\circ$ and $4^\circ$ of rotation error).

\paragraph{Results.} \cref{tab:ac-yield} reports the yield. The Point column is nearly identical for \ours-B/L/G and \blomaff, which use the matches of their LoMa backbones (G for \blomaff), so the comparison against \blomaff~isolates the affine head: \blomaff~loses most of its yield to the affine criterion (from \num{29.1}\% to \num{2.8}\% at $\tau_A=0.1$ on Mega1500), whereas \ours-G loses little (to \num{23.0}\%). \sunetal~has a higher point recall on Mega1500 (\num{57.1}\%), but a lower yield at $\tau_A=0.25$ than ours (\num{17.4}\% against \num{27.3}\%).

\begin{table}[t]
\centering
\caption{\textbf{End-to-end AC yield.} Percentage of a budget of \num{2048} keypoints that yields an AC with a correct point ($\tau_p=\num{3}\,$px) and affine (\cref{eq:ac-correct}). The columns $0.25$ and $0.1$ give $\tau_A$. \emph{Point} drops the affine criterion. Higher is better.}
\label{tab:ac-yield}
\small

\setlength{\tabcolsep}{3pt}

\begin{tabular}{l@{\hspace{6pt}}rrr@{\hspace{10pt}}rrr}
\toprule
Method &
\multicolumn{3}{c}{HPatches} &
\multicolumn{3}{c}{Mega1500} \\
\cmidrule(lr){2-4}
\cmidrule(lr){5-7}
& Point & $0.25$ & $0.1$
& Point & $0.25$ & $0.1$ \\
\midrule

\refrow{RoMa v2} & \refrow{96.4} & \refrow{94.5} & \refrow{89.3} & \refrow{72.0} & \refrow{63.1} & \refrow{47.8} \\
\midrule
\baffnet & 39.1 & 17.7 & 2.8 & 8.7 & 3.0 & 0.5 \\
\bsesti &  \best 75.0 & 13.2 & 2.0 & 12.7 & 3.0 & 0.7 \\
\sunetal &  66.5 & 11.8 & 2.0 & \best 57.1 & 17.4 & 4.9 \\
\blomaff & 51.2 & 9.8 & 1.7 & 29.1 & 9.1 & 2.8 \\
\ours-B & 51.1 & 50.9 & 50.1 & 29.0 & 27.0 & 22.5 \\
\ours-L & 51.1 & 50.9 & 49.8 & 29.1 & 27.2 & 22.8 \\
\ours-G &  51.2 & \best 51.0 & \best 50.3 & 29.1 & \best 27.3 & \best 23.0 \\

\bottomrule
\end{tabular}
\end{table}

\subsection{Affine Accuracy in the Pixel Frame}\label{subsec:pixel-frame}

\ours~is trained in the normalized frame $[-1,1]^2$, whereas the patch-based baselines operate in pixel space. \cref{tab:affine-frob-pix} repeats \cref{tab:affine-frob} in the pixel frame, $\mat{A}^{\mathrm{pix}}=\mat{S}_{\mathcal B}\mat{A}\mat{S}_{\mathcal A}^{-1}$ with $\mat{S}=\operatorname{diag}(W/2,H/2)$ of each image at its native resolution, on the same matches. The residual rotation is read off $\hat{\mat{A}}^{\mathrm{pix}}(\mat{A}^{\mathrm{pix}})^{-1}$ in the same frame. The two frames agree within about \num{6}\% for every method and the ranking is unchanged.

\begin{table}[t]
\centering
\caption{\textbf{Accuracy of affine transformations in the pixel frame.} Median Frobenius ($\times100$) and rotational errors on the matches of \cref{tab:affine-frob} (lower is better).}
\label{tab:affine-frob-pix}
\small

\setlength{\tabcolsep}{3pt}

\begin{tabular}{l@{\hspace{6pt}}rr rr}
\toprule
Method &
\multicolumn{2}{c}{HPatches} &
\multicolumn{2}{c}{Mega1500} \\
\cmidrule(lr){2-3}
\cmidrule(lr){4-5}
& Frobenius & Rot. & Frobenius & Rot. \\
\midrule
\baffnet & 25.8 & 6.03$^\circ$ & 34.3 & 7.66$^\circ$ \\
\bsesti & 51.2 & 8.83$^\circ$ & 48.8 & 5.24$^\circ$ \\
\sunetal & 47.3 & 3.39$^\circ$ & 40.1 & 2.00$^\circ$ \\
\blomaff & 50.0 & 4.46$^\circ$ & 42.9 & 2.40$^\circ$ \\
\ours-B & 1.5 & 0.25$^\circ$ & 4.1 & 0.61$^\circ$ \\
\ours-L & 1.6 & 0.24$^\circ$ & 4.0 & 0.60$^\circ$ \\
\ours-G & \best 1.5 & \best 0.24$^\circ$ & \best 3.8 & \best 0.57$^\circ$ \\
\bottomrule
\end{tabular}
\end{table}

\subsection{Sensitivity to the Ground-Truth Construction}\label{subsec:gt-sensitivity}

The supervision of \cref{subsec:supervision} is a finite-difference Jacobian (\cref{eq:findiff}) of the depth-induced warp, a first-order approximation. The HPatches GT, the closed-form Jacobian of a known homography, uses no depth, no finite differences and no masking, yet \ours~is the most accurate method there by an order of magnitude (\cref{tab:affine-frob}). We also measure the truncation error of the step directly: on HPatches, \cref{tab:gt-sensitivity} gives the Frobenius distance between the forward difference and the analytic Jacobian as a function of the step $h$. The error grows linearly in $h$, and at our step $h=1$\,px it is $0.03$ ($\times100$, median), about $55\times$ below the $1.50$ error of \ours-G on the same data.
\begin{table}[t]
\centering
\caption{\textbf{Finite-difference step vs.\ the exact Jacobian on HPatches.} Frobenius distance ($\times100$) between the forward difference with step $h$ (px) and the analytic Jacobian. Training uses $h=1$\,px.}
\label{tab:gt-sensitivity}
\small
\setlength{\tabcolsep}{4pt}
\begin{tabular}{lrrrrrr}
\toprule
$h$ (px) & 0.25 & 0.5 & 1 & 2 & 4 & 8 \\
\midrule
Median & 0.007 & 0.014 & 0.027 & 0.054 & 0.108 & 0.217 \\
Mean & 0.010 & 0.021 & 0.042 & 0.083 & 0.167 & 0.334 \\
95th pct. & 0.031 & 0.062 & 0.124 & 0.248 & 0.496 & 0.993 \\
\bottomrule
\end{tabular}
\end{table}

\subsection{Full Orientation Estimation}\label{supp:rotation}

Prior work based on local patches has found it difficult to predict orientation, scale and affine shape jointly~\cite{mishkin2018repeatability, sun2025learning}. Here we show that our setup predicts full affine frames, including orientation. We compare the LoMa-B matcher with the rotation-invariant LoMa-R matcher~\cite{nordstrom2026whohandlesorientation}, which has the same architecture, on independently rotated image pairs, following~\cite{nordstrom2026whohandlesorientation}. \cref{tab:rotation} has four rows: frozen LoMa-B without rotation augmentation (the upright baseline), frozen LoMa-R, unfrozen LoMa-R, and unfrozen LoMa-B initialized from \ours-B, the last two being the proposed variants. Frozen LoMa-R gives subpar results, even compared to the upright baseline, because its features are rotation invariant and cannot be used to estimate orientation, although it matches rotated images better than LoMa-B. The two proposed variants give both good affine estimates and good matches and are comparable, so the LoMa-B feature dimension suffices for both rotation-robust matching and orientation estimation.

\begin{table}
\centering
\caption{\textbf{Results under rotation.}  \snow~and \flame~denote using a frozen and un-frozen matcher, respectively. RA is short for rotation augmentation during training. We denote the upright evaluation by $\uparrow$ and the rotated by $\circlearrowright$.}
\label{tab:rotation}

\setlength{\tabcolsep}{3pt}
\begin{tabular}{l@{\hspace{6pt}}cc rrrr rr}
\toprule
& & & \multicolumn{2}{c}{HPatches} & \multicolumn{4}{c}{Mega1500} \\
\cmidrule(lr){4-5} \cmidrule(lr){6-9}
& & & \multicolumn{2}{c}{Aff. Frob.} & \multicolumn{2}{c}{Aff. Frob.} & \multicolumn{2}{c}{AUC@$20^\circ$} \\
\cmidrule(lr){4-5} \cmidrule(lr){6-7} \cmidrule(lr){8-9}
Matcher & \snow/\flame & RA & $\uparrow$ & $\circlearrowright$ & $\uparrow$ & $\circlearrowright$ & $\uparrow$ & $\circlearrowright$ \\
\midrule
LoMa-B & \snow & \xmark & 1.62 & 3.30 & \bfseries 3.97 & 7.01 & \bfseries 83.6 & 81.2 \\
\dashedmidrule
LoMa-R & \snow & \cmark & 8.00 & 9.01 & 8.74 & 8.96 & 82.8 & \bfseries 83.2 \\
\dashedmidrule
LoMa-R & \flame & \cmark & 1.69 & 1.83 & 4.06 & \bfseries 4.11 & 83.4 & 82.8 \\
LoMa-B & \flame & \cmark & \bfseries 1.58 & \bfseries 1.70 & 4.27 & 4.58 & 82.8 & 82.6 \\
\bottomrule
\end{tabular}

\end{table}

\section{Gauge Freedom}\label{supp:gauge}

The loss of \cref{eq:loss} constrains only the composition $\hat{\mat{A}}_{ij}=\hat{\mat{A}}^{\mathcal B}_j(\hat{\mat{A}}^{\mathcal A}_i)^{-1}$, so the per-keypoint frames are determined only up to a common right factor $\mat{G}$ (\cref{fig:gauge}(a)). How much of this freedom the trained model uses is an empirical question.

\paragraph{Protocol.} Using \ours-G on HPatches viewpoint sequences, we fix the first image of a sequence and pair it with $P=5$ other views, which gives $P$ predictions $\hat{\mat{A}}_1,\dots,\hat{\mat{A}}_P$ of the frame of each of its keypoints. For every keypoint and pair of partners $(p,q)$ we measure the residual gauge
\begin{equation}
    \mat{G}_{pq} = \big(\hat{\mat{A}}_p\big)^{-1}\hat{\mat{A}}_q,
    \label{eq:residual-gauge}
\end{equation}
which is the identity exactly when both partners induce the same frame, and summarize it by the median of $\|\mat{G}_{pq}-\mat{I}\|_F$ over keypoints and partner pairs. Even for stable frames two estimates differ by their estimation error, so we compare with the affine error $\epsilon$ of \cref{tab:affine-frob} on the same data ($1.50$ for \ours-G on HPatches, $\times100$), which gives an expected distance of roughly $\sqrt{2}\,\epsilon$ to $2\epsilon$, \ie, $2.1$ to $3.0$.

\paragraph{Results.} \cref{fig:gauge}(b,c) shows the frames of one image under five partners and the drift over six HPatches viewpoint sequences. The median drift ($\times100$) is $30$ (\emph{london}), $33$ (\emph{coffeehouse}), $68$ (\emph{tempera}), $80$ (\emph{wormhole}), $84$ (\emph{calder}) and $130$ (\emph{gardens}), an order of magnitude or more above the $2.1$ to $3.0$ expected for stable frames. The gauge is therefore largely unresolved. This means that a frame predicted for one pair is not a view-independent quantity, which is what the transport (\cref{supp:transport}) aims to address.

\section{Per-View Frames by Transport}\label{supp:transport}

\begin{figure}[t]
\centering
\includegraphics[width=0.99\linewidth]{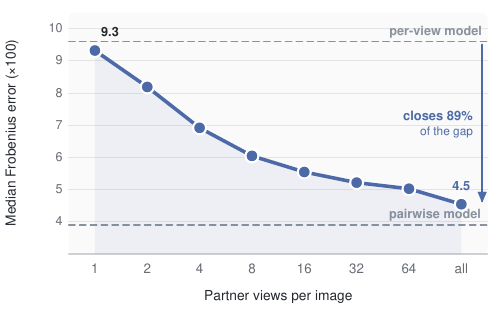}
\caption{\textbf{Averaged frames improve with the number of partners.} Affine error (Frobenius) on MegaDepth-1500 against the number of partner views $N$, from the per-view head ($N=0$) to the pairwise \ours-B.}
\label{fig:partners}
\end{figure}

The frames $\hat{\mat{A}}^{\mathcal A}_i$ and $\hat{\mat{A}}^{\mathcal B}_j$ of a pair-conditioned \ours~depend on the partner image (\cref{supp:gauge}), whereas applications, such as SfM, benefit from a frame per keypoint that is independent of any other view. A head that reads a single image gives such a frame, $\tilde{\mat{A}}_i^{\mathcal P}$ for keypoint $\mat{x}_i^{\mathcal P}$ in image $\mat{I}^{\mathcal P}$, but is less accurate than the relative affine $\hat{\mat{A}}_{ij}$ of the conditioned model. We combine the two by \emph{transport}. Let $\mat{I}^{\mathcal T}$ be the image whose frames we want and $\mat{I}^{\mathcal P}$ another view of the scene, with a match $(i,j)$ between $\mat{x}_i^{\mathcal P}$ and $\mat{x}_j^{\mathcal T}$. The relative affine $\hat{\mat{A}}_{ij}$ carries the per-view frame of $\mat{x}_i^{\mathcal P}$ to
\begin{equation}
    \hat{\mat{A}}_j^{\mathcal T \leftarrow \mathcal P}
    \;=\;
    \hat{\mat{A}}_{ij}\,\tilde{\mat{A}}_i^{\mathcal P},
    \label{eq:transport}
\end{equation}
an estimate of the frame of $\mat{x}_j^{\mathcal T}$ in the convention of the per-view head. We average \cref{eq:transport} over $N$ partner views in the parametrization of the head (\cref{eq:decomp}). Each transported matrix is decomposed into scale, rotation and shape as in \cref{eq:decomp}, then the log-scale, log shape exponent and shear are averaged arithmetically and the orientation as a circular mean (the normalized mean of $(\cos\alpha,\sin\alpha)$). A keypoint that no partner supports keeps $\tilde{\mat{A}}_j^{\mathcal T}$. On MegaDepth-1500 we use up to $N$ random views of the same scene with at least 100 matches to $\mat{I}^{\mathcal T}$ (never the two images of the evaluated pair), and \emph{all} uses every other view. The relative affines come from \ours-B. The errors of the per-view head in different views are roughly independent and average down, while the accurate relative affines carry each frame between views. With $N=0$ this is the per-view head, and \cref{fig:partners} shows how the error decreases with $N$. Averaging roughly halves the error of the single-view head. \citet{eichhardt2019optimalmultiview} similarly correct local affine frames using multiple views, though geometrically.

\section{More Qualitative Results}

\begin{figure*}[t]
\centering
\includegraphics[width=0.99\linewidth]{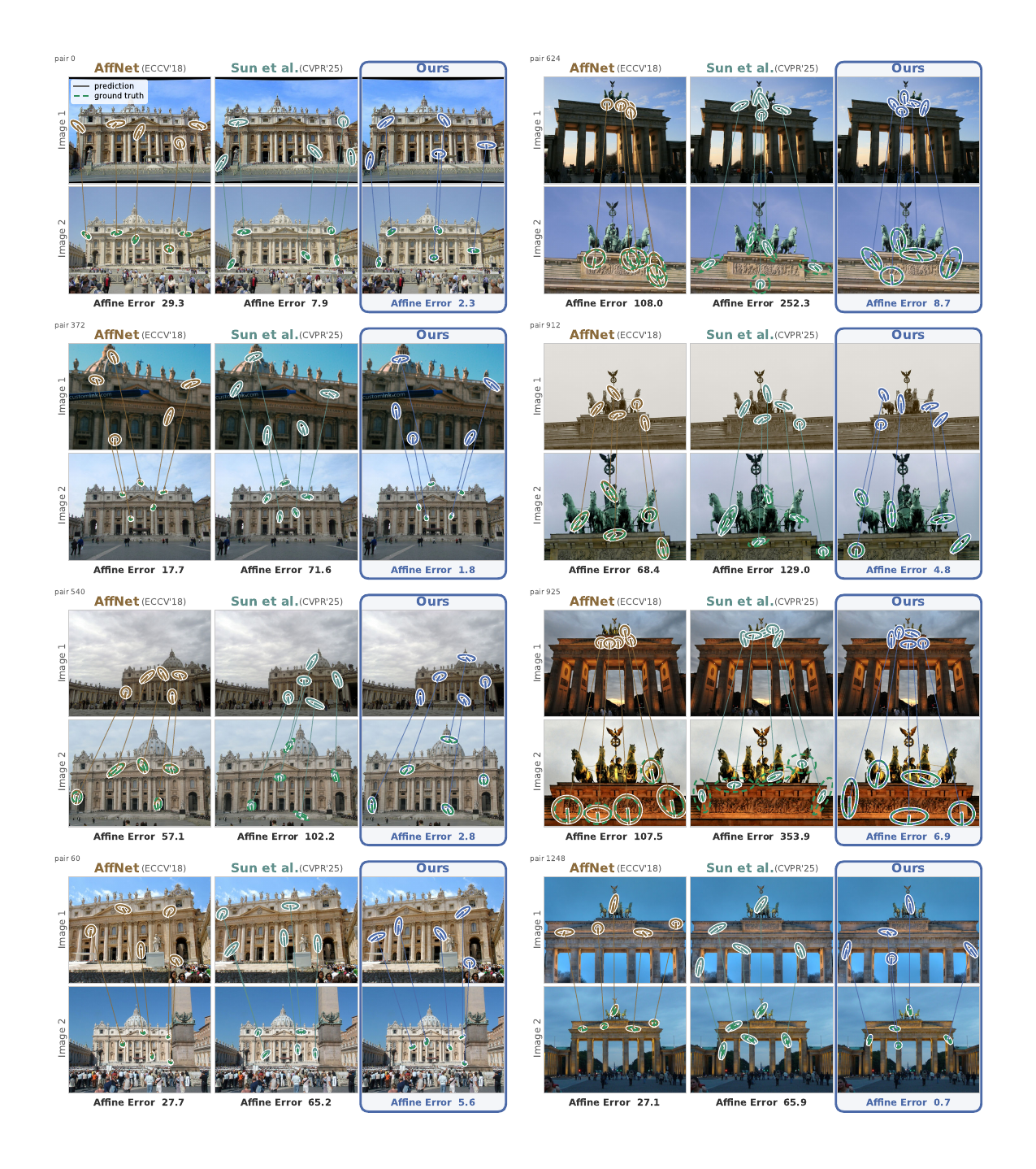}
\caption{\textbf{Visualizing further affine estimations.} Pairs from the two MegaDepth-1500 scenes, 0015 (left) and 0022 (right). Errors are median Frobenius ($\times100$).}
\label{fig:qualitative-supp}
\end{figure*}

In \cref{fig:qualitative-supp}, we provide additional examples of AC estimation on MegaDepth-1500. We compare \ours-B to KN+HN+AN (AffNet) and \sunetal.

\end{document}